\documentclass{article}

\usepackage{PRIMEarxiv}

\usepackage[utf8]{inputenc} 
\usepackage[T1]{fontenc}    
\usepackage{hyperref}       
\usepackage{url}            
\usepackage{booktabs}       
\usepackage{amsfonts}       
\usepackage{fancyhdr}       
\usepackage{graphicx}       

\usepackage[numbers,sort&compress]{natbib}
\usepackage{amsmath}
\usepackage{amssymb}
\usepackage{amsthm}
\usepackage{caption}
\usepackage{enumitem}
\usepackage{xcolor}
\usepackage{geometry}
\usepackage{tikz}
\usetikzlibrary{shapes,arrows,positioning,calc,fit,shapes.geometric,arrows.meta}
\usepackage{listings}
\usepackage{upquote}
\usepackage{setspace}

\DeclareUnicodeCharacter{2200}{$\forall$}
\DeclareUnicodeCharacter{2203}{$\exists$}
\DeclareUnicodeCharacter{2227}{$\wedge$}
\DeclareUnicodeCharacter{2227}{$\wedge$}
\DeclareUnicodeCharacter{2223}{$\mid$}
\DeclareUnicodeCharacter{2264}{$\leq$}
\DeclareUnicodeCharacter{2115}{$\mathbb{N}$}
\DeclareUnicodeCharacter{220F}{$\Pi$}
\DeclareUnicodeCharacter{2265}{$\geq$}
\DeclareUnicodeCharacter{00D7}{$\times$}
\DeclareUnicodeCharacter{2014}{---}

\lstdefinestyle{mintedstyle}{
    backgroundcolor=\color{gray!5},
    frame=lines,
    framesep=2mm,
    xleftmargin=3mm,
    xrightmargin=3mm,
    framexleftmargin=3mm,
    framexrightmargin=3mm,
    breaklines=true,
    basicstyle=\ttfamily\small\setstretch{1.1},
    keywordstyle=\color{blue!70!black}\bfseries,
    commentstyle=\color{green!50!black},
    stringstyle=\color{orange!90!black},
    showstringspaces=false,
    columns=fullflexible,
    mathescape=true,
    literate={
        {→}{{$\to$ }}2
        {∀}{{$\forall$ }}1
        {∃}{{$\exists$ }}1
        {ℕ}{{$\mathbb{N}$}}1
        {ℝ}{{$\mathbb{R}$}}1
        {ℤ}{{$\mathbb{Z}$}}1
        {≤}{{$\le$ }}1
        {≥}{{$\ge$ }}1
        {∧}{{$\wedge$ }}1
        {∨}{{$\vee$}}1
        {¬}{{$\neg$}}1
        {∏}{{$\prod$ }}1
        {×}{{$\times$ }}1
        {∣}{{$\mid$ }}1
    }
}

\lstdefinelanguage{lean}{
    classoffset = 1,
    morekeywords={
        import,theorem,set_option,lemma,def,example,by,have,calc,fun,match,with,if,then,else,do,
        let,in,Type,where,inductive,structure,deriving,mutual,namespace,
        end,open,variable,variables,section,universe,using,pattern,macro,
        notation,macro_rules,axiom,constant
    },
    keywordstyle=\color{blue!70!black}\bfseries,
    classoffset = 2,
    morekeywords={
        Prop,sorry
    },
    keywordstyle=\color{red!70!black}\bfseries,
    sensitive=true,
    morecomment=[l]{--},
    morecomment=[s]{/-}{-/},
    morestring=[b]",
    morestring=[s]{`}{`},
    upquote=true
}

\lstdefinelanguage{markdown}{
    morestring=[b]",
    morestring=[s]{```}{```},
    morecomment=[l]{\%},
    stringstyle=\color{black},
    identifierstyle=\color{black},
    keywordstyle=\color{black}
}

\lstdefinelanguage{ini}{
    morecomment=[l]{;},
    morecomment=[s][\color{green!40!black}]{[}{]},
    morestring=[b]"
}

\lstdefinelanguage{bash}{
    morekeywords={if,then,else,fi,for,in,do,done,echo,exit,while,case,esac,elif,function},
    sensitive=true,
    morecomment=[l]{\#},
    morestring=[b]"
}

\lstnewenvironment{minted}[2][]{%
    \lstset{style=mintedstyle,language=#2,#1}%
}{}

\title{Gödel's Poetry}

\author{
  Kelly J. Davis \\
  Unaffiliated \\
  \texttt{kdavis@alum.mit.edu}
}

\begin{document}
\maketitle

\begin{abstract}
Formal, automated theorem proving has long been viewed as a challenge to artificial intelligence. We introduce here a new approach to computer theorem proving, one that employs specialized language models for Lean4 proof generation combined with recursive decomposition of difficult theorems into simpler entailing propositions. These models are coordinated through a multi-agent architecture that orchestrates autoformalization (if required), proof generation, decomposition of difficult theorems into simpler entailing propositions, and recursive proof (and/or decomposition) of these propositions. Without decomposition, we achieve a 90.4\% pass rate on miniF2F. With decomposition, this is significantly improved. A key technical contribution lies in our extension of the Kimina Lean Server with abstract syntax tree (AST) parsing capabilities to facilitate automated, recursive proof decomposition. The system is made available on PyPI as \texttt{goedels-poetry}, and the open-source implementation at \url{https://github.com/KellyJDavis/goedels-poetry} facilitates both adaptation to alternative language models and extension with custom functionality.
\end{abstract}

\section{Introduction}

Formal, automated theorem proving represents a fundamental challenge to artificial intelligence. The task requires automated generation of formal proofs that may be verified by computer systems~\citep{harrison2014history}. Recent advances in large language models~\citep{xai2025grock4,openai2025gpt5,anthropic2025claudesonnet4.5,anthropic2025claudehaiku4.5} have demonstrated remarkable capabilities in mathematical reasoning~\citep{goedel-prover-v2,ren2025deepseek,wang2025kimina,hilbert,chen2025seedproverdeepbroadreasoning,zhou2025solvingformalmathproblems,xin2025scalingmultiturnoffpolicyrl,achim2025aristotleimolevelautomatedtheorem}. However, the generation of formally verified proofs remains a difficult undertaking due to the strict syntactic and logical requirements imposed by proof assistants such as Lean~\citep{lean4}, Isabelle~\citep{isabelle}, and Coq~\citep{coq}.

We introduce here Gödel's Poetry, a system combining automated theorem proving, decomposition of difficult theorems into entailing propositions, RAG based retrieval of propositions useful for decomposition, and specialized language models. Our approach builds upon three recent advances: the verifier-guided self-correction approach of Goedel-Prover-V2~\citep{goedel-prover-v2}, the recursive proof strategy of POETRY~\citep{poetry}, and the RAG based propositions retrieval of Hilbert~\citep{hilbert}. The system coordinates specialized agents through LangGraph~\citep{langgraph} and LangChain~\citep{langchain} for the purposes of autoformalization, autoformalization verification, proof generation, proof verification, and recursive decomposition.

A key technical contribution of this work lies in our extension of the Kimina Lean Server~\citep{kimina} to support abstract syntax tree (AST) extraction from Lean 4 code. This extension enables programmatic analysis of decompositions, automatic identification of unproven subgoals, and extraction of subgoal proposition statements—operations that prove essential for recursive decomposition as adapted to Lean's tactic-based proof system.

The system provides three principal capabilities: (1) multi-stage proof generation proceeding through an optional autoformalization phase (autoformalization, autoformalization syntactic validation, and autoformalization semantic validation), a proof generation phase , and a proof verification phase; (2) recursive decomposition of complex theorems into entailing propositions using proof sketches with \texttt{sorry} placeholders; (3) proof reconstruction integrating verified proposition proofs through AST-based substitution. The modular architecture permits the substitution of alternative language models, thereby facilitating experimentation with different proof strategies.

\section{Related Work}

\subsection{Neural Theorem Proving}

Neural theorem proving, which combines deep learning with formal verification, has emerged as a productive area of research~\citep{li2024surveydeeplearningtheorem}. Its modern era arguably began with Polu and Sutskever's work~\citep{polu2020generative} applying transformer-based~\citep{vaswani2023attentionneed} language models to automated theorem proving. Formulating tactic prediction as language modeling, this inspired much follow-on research~\citep{polu2023formal, han2022proof, jiang2021lisa, zhang2023learning, yeh2023coprover, xiong2023trigo, welleck2021naturalproofs, vishwakarma2023enhancing}.

More recently, the work of DeepSeek-Prover~\citep{xin2024deepseek} demonstrated that language models, when fine-tuned on mathematical corpora, are capable of generating Lean 4 proofs with significant success rates. This motivated follow-on work, DeepSeek-Prover-V1.5~\citep{huajian2024deepseekproverv15} and DeepSeek-Prover-V2~\citep{ren2025deepseek}. Subsequent to DeepSeek-Prover, AlphaProof~\citep{trinh2024alphaproof} achieved silver-medal performance on the International Mathematical Olympiad employing reinforcement learning with formal verification. Even more recently there has been a flood of papers from Apple~\citep{hilbert}, ByteDance~\citep{zhou2025solvingformalmathproblems, xin2025scalingmultiturnoffpolicyrl, chen2025seedproverdeepbroadreasoning}, Harmonic~\citep{achim2025aristotleimolevelautomatedtheorem}, Kimina~\citep{wang2025kimina}, and many others.

Progress in this domain has been driven by benchmarks including miniF2F~\citep{zheng2021minif2f}, MathLib-Bench~\citep{song2024mathlib}, and PutnamBench~\citep{lu2024putnam}, which provide standardized evaluation on undergraduate mathematics, Mathlib~\citep{mathlib} theorems, and Putnam competition problems.

\subsection{Goedel-Prover-V2}

The work of Goedel-Prover-V2~\citep{goedel-prover-v2} introduced three key innovations: (1) scaffolded data synthesis, which generates training examples of increasing difficulty; (2) verifier-guided self-correction, which enables iterative proof refinement using compiler feedback; and (3) model averaging, which serves to mitigate diversity loss during training. Their 8B model achieved 84.6\% pass@32 on miniF2F, thereby outperforming DeepSeek-Prover-V2-671B despite being 80× smaller. Their 32B model reached 88.1\% on miniF2F at pass@32 in standard mode and 90.4\% at pass@32 with verifier-guided self-correction. On PutnamBench, the model solved 86 problems at pass@184, at the time securing first place among open-source models.

Our system employs Goedel-Prover-V2 models that use verifier-guided self-correction as specialized agents for direct proof generation. This is then extended through the addition of RAG-guided, hierarchical decomposition for cases in which a direct proof fails.

\subsection{POETRY: Proving Theorems Recursively}

The work of POETRY~\citep{poetry} introduced recursive, level-by-level proof generation for Isabelle~\citep{isabelle}. Rather than generating proofs in a step-by-step fashion, POETRY creates verifiable sketches, i.e. decompositions, at each level, employing \texttt{sorry} placeholders for intermediate propositions. The system focuses upon solving the theorem at the current level, thereby deferring detailed proofs to subsequent levels. This approach has been found to discover substantially longer proofs—increasing the maximum proof length from 10 to 26 steps—and achieved a 5.1\% improvement on miniF2F.

We adapt this recursive strategy, implementing proof decompositions wherein difficult theorems are decomposed into entailing propositions with explicit named hypotheses (expressed as \texttt{have} statements). These \texttt{have} statements are then directly proven. If direct proof is too difficult, then they too are sketched and decomposed into subgoals with explicit named hypotheses. This entire process is applied recursively.

\subsection{Hilbert: Retrieving Relevant Theorems}

Concurrently with our work, the Hilbert system~\citep{hilbert} demonstrated that recursive decomposition was aided by retrieving previously proven theorems that could aid in decomposition. When tasked with decomposing a theorem, an LLM in the Hilbert system~\citep{hilbert} system generates a number descriptions of theorems it thinks would be of use in decomposition. These descriptions are then used to query a vector database to retrieve relevant theorems. These theorems are then, if appropriate, employed in decomposition. This framework yields substantial improvements on formal theorem proving benchmarks, achieving 99.2\% on miniF2F and 70.0\% on PutnamBench. While Hilbert was developed independently, the convergence upon these recursive decomposition strategies serves to validate this architectural approach.

We adopt this RAG-bsed theorem retrieval. G\"odel's Poetry adds an agent that generates a number of descriptions of theorems it thinks would be of use in decomposition. It then queries a vector database, we employ a LeanExplore variant~\citep{asher2025leanexploresearchenginelean}, to retrieve relevant theorems that are, if appropriate, then used in decomposition.

\subsection{Kimina Lean Server}

The Kimina Lean Server~\citep{kimina} provides high-performance parallel verification of Lean 4 proofs through a FastAPI~\citep{fastapi} service that wraps the Lean REPL. It supports batch proof checking with configurable parallelism, REPL reuse for improved throughput, and structured error reporting.

We extend the Kimina server with AST extraction capabilities (the details of which appear in Appendix~\ref{sec:appendix-kimina}), thereby enabling programmatic analysis of proof structures essential for recursive decomposition.

\subsection{Multi-Agent Systems}

Multi-agent architectures have been applied with success to a variety of AI tasks~\citep{xu2025comprehensivesurveydeepresearch}. In the domain of theorem proving, HyperTree Proof Search~\citep{lample2022hypertree} employs a MCTS-inspired search algorithm. The work of Draft-Sketch-Prove~\citep{jiang2023draft} uses separate models for drafting informal proofs, sketching formal structure, and completing details. Hilbert~\citep{hilbert} introduces four agents: an informal LLM that excels at mathematical reasoning, a specialized prover LLM optimized for Lean 4 tactics, a formal verifier, and a semantic theorem retriever.

Our system employs specialized agents for distinct stages of the proof process (e.g. formalization, proving, decomposition), with orchestration through LangGraph~\citep{langgraph} enabling flexible workflow management.

\section{Methods}

\subsection{System Architecture}

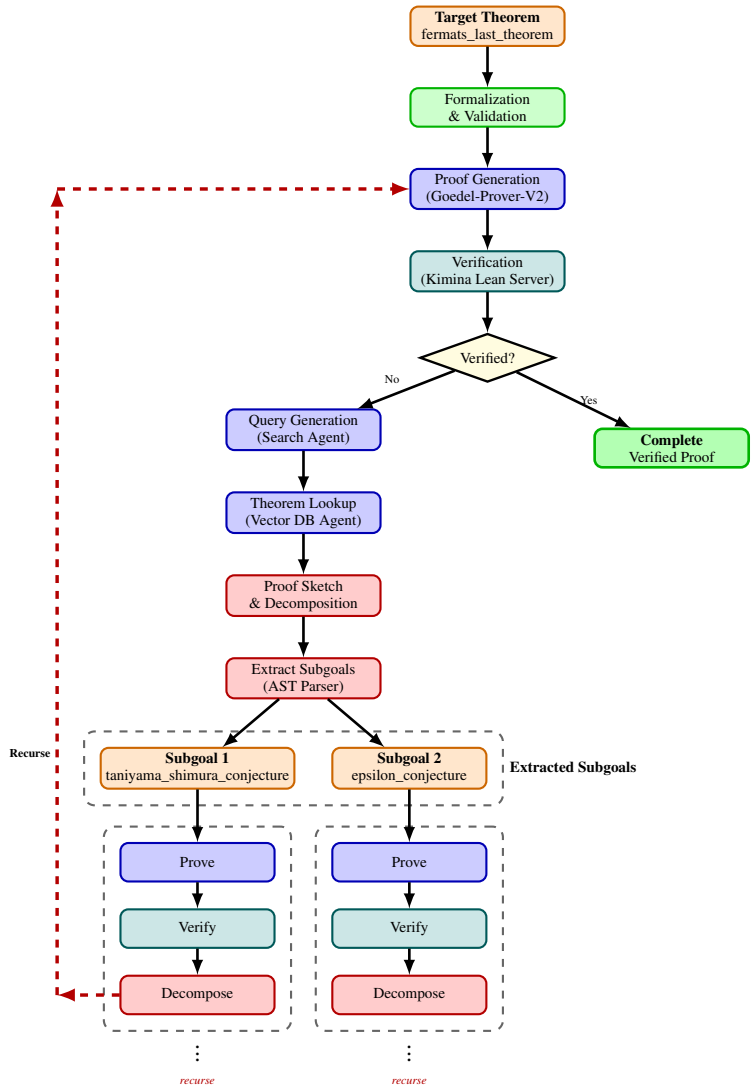
\begin{figure}[htbp]
\centering
\begin{tikzpicture}[scale=0.64, transform shape,
    node distance=0.85cm and 2.0cm,
    box/.style={draw, rounded corners=3pt, minimum width=3.2cm, minimum height=0.8cm, align=center, font=\small, inner sep=3pt},
    theorem/.style={box, fill=orange!20, draw=orange!80!black, thick},
    prover/.style={box, fill=blue!20, draw=blue!70!black, thick},
    verify/.style={box, fill=teal!20, draw=teal!70!black, thick},
    decompose/.style={box, fill=red!20, draw=red!70!black, thick},
    formalize/.style={box, fill=green!20, draw=green!70!black, thick},
    decision/.style={diamond, draw=black, line width=0.9pt, aspect=2.5, minimum width=2.8cm, align=center, fill=yellow!15, font=\small},
    output/.style={box, fill=green!30, draw=green!70!black, line width=1pt},
    dashedbox/.style={draw=black!60, thick, dashed, inner sep=6pt, rounded corners=4pt},
    arrow/.style={-latex, line width=1.0pt, draw=black},
    recursearrow/.style={-latex, line width=1.3pt, draw=red!70!black, dashed}
]

\node[theorem] (target) {\textbf{Target Theorem}\\fermats\_last\_theorem};
\node[formalize, below=of target] (formalize) {Formalization\\  \& Validation};
\node[prover, below=of formalize] (prover) {Proof Generation\\(Goedel-Prover-V2)};
\node[verify, below=of prover] (verify) {Verification\\(Kimina Lean Server)};
\node[decision, below=of verify] (success) {Verified?};
\node[output, below right=1.2cm and 1.5cm of success] (complete) {\textbf{Complete}\\Verified Proof};

\node[prover, below left=0.8cm and 1.5cm of success] (querygen) {Query Generation\\(Search Agent)};
\node[prover, below=of querygen] (theoremlookup) {Theorem Lookup\\(Vector DB Agent)};
\node[decompose, below=of theoremlookup] (sketch) {Proof Sketch\\  \& Decomposition};
\node[decompose, below=of sketch] (extract) {Extract Subgoals\\(AST Parser)};

\node[theorem, below=1.0cm of extract, xshift=-2.2cm] (sg1) {\textbf{Subgoal 1}\\taniyama\_shimura\_conjecture};
\node[theorem, below=1.0cm of extract, xshift=2.2cm] (sg2) {\textbf{Subgoal 2}\\epsilon\_conjecture};

\node[dashedbox, fit=(sg1) (sg2), label={[font=\small\bfseries]right:Extracted Subgoals}] (subgoals) {};

\node[prover, below=1.1cm of sg1] (rp1) {Prove};
\node[verify, below=0.55cm of rp1] (rv1) {Verify};
\node[decompose, below=0.55cm of rv1] (rd1) {Decompose};

\node[prover, below=1.1cm of sg2] (rp2) {Prove};
\node[verify, below=0.55cm of rp2] (rv2) {Verify};
\node[decompose, below=0.55cm of rv2] (rd2) {Decompose};

\node[dashedbox, fit=(rp1) (rv1) (rd1)] (rec1) {};
\node[dashedbox, fit=(rp2) (rv2) (rd2)] (rec2) {};

\node[below=0.3cm of rd1, font=\Large] (dots1) {$\vdots$};
\node[below=0.3cm of rd2, font=\Large] (dots2) {$\vdots$};
\node[below=0.1cm of dots1, font=\scriptsize, text=red!70!black] {\textit{recurse}};
\node[below=0.1cm of dots2, font=\scriptsize, text=red!70!black] {\textit{recurse}};

\draw[arrow] (target) -- (formalize);
\draw[arrow] (formalize) -- (prover);
\draw[arrow] (prover) -- (verify);
\draw[arrow] (verify) -- (success);
\draw[arrow] (success) -- node[right, font=\scriptsize] {Yes} (complete);
\draw[arrow] (success) -- node[above left, font=\scriptsize] {No} (querygen);
\draw[arrow] (querygen) -- (theoremlookup);
\draw[arrow] (theoremlookup) -- (sketch);
\draw[arrow] (sketch) -- (extract);
\draw[arrow] (extract) -- (sg1);
\draw[arrow] (extract) -- (sg2);
\draw[arrow] (sg1) -- (rp1);
\draw[arrow] (rp1) -- (rv1);
\draw[arrow] (rv1) -- (rd1);
\draw[arrow] (sg2) -- (rp2);
\draw[arrow] (rp2) -- (rv2);
\draw[arrow] (rv2) -- (rd2);

\draw[recursearrow] (rd1.west) -- ++(-1.3,0) coordinate (rec_left);
\draw[recursearrow] (rec_left) -- node[left, font=\scriptsize, pos=0.3] {\textbf{Recurse}} (rec_left |- prover.west) coordinate (rec_mid);
\draw[recursearrow] (rec_mid) -- (prover.west);

\end{tikzpicture}
\caption{\textbf{The Gödel's Poetry recursive decomposition architecture.} Flow proceeds top-to-bottom: a target theorem undergoes (if required) formalization, proof generation with Goedel-Prover-V2 using verifier-guided self-correction, and verification with Kimina Lean Server. Upon verification failure, the system queries a vector database, obtaining theorems of use when generating a proof sketch. Using these theorems it generates a proof sketch with \texttt{sorry} placeholders, extracts subgoals via AST parsing, and recursively proves each (shown in dashed boxes with colored stages: orange=theorem, blue=prove, teal=verify, red=decompose). The red dashed arrow shows recursion back to the prover for subgoal solving.}
\label{fig:architecture}
\end{figure}

Our system employs a multi-agent architecture  (Figure~\ref{fig:architecture}) that is coordinated by a supervisor agent, elided in (Figure~\ref{fig:architecture}). The supervisor analyzes the current proof state and dispatches work to specialized agents, shown in (Figure~\ref{fig:architecture}), which in turn handle formalization and validation of formalized theorems, proof generation, proof validation, query generation, theorem lookup, proof sketch/decomposition, and extraction of subgoals.

The system maintains a tree-structured proof state, wherein each node represents either a leaf node (containing the formal theorem, proof attempt, verification status, and proof history) or an internal node (containing the formal theorem, proof decomposition, child subgoals, and decomposition history). This structure enables both the tracking of complete proof provenance and the systematic reconstruction of verified proofs.

\subsection{Formalization Pipeline}

For informal theorems expressed in natural language, the system executes a three-stage pipeline, which we now describe.

\textbf{Stage 1: Formalization.} The formalizer agent employs a swappable LLM, which by default is Goedel-Formalizer-V2, to translate a natural language statement into a Lean 4 theorem (the prompt it employs is provided in full in Appendix~\ref{sec:appendix-formalizer-prompt}). The model was fine-tuned upon pairs of informal mathematical statements and their formal Lean 4 counterparts.

\textbf{Stage 2: Syntax Validation.} The syntax agent submits the formalization to the Kimina Lean Server~\citep{kimina} for parsing. The Lean compiler reports any syntax errors, type mismatches, unknown identifiers, or inconsistencies. Invalid formalizations are returned to Stage 1, and this process continues until validation succeeds or the maximum number of retries is exhausted (default: 10 attempts).

\textbf{Stage 3: Semantic Checking.} A semantic validation agent employs a swappable LLM, which is by default Qwen 3 30GB~\citep{qwen2025}, to verify that the formalization preserves the meaning of the informal statement (the prompt it employs is provided in full in Appendix~\ref{sec:appendix-semantic-prompt}). The agent receives both the original informal statement and the formalized version, and determines whether the formalization accurately captures the meaning of the informal statement. This serves to prevent syntactically valid but semantically incorrect translations. The semantic checker possesses veto power: if it determines that the formalization is incorrect, the theorem is returned to Stage 1.

\subsection{Direct Proof Generation}

The prover agent employs a swappable LLM, which is by default the Goedel-Prover-V2~\citep{goedel-prover-v2} model, and uses verifier-guided self-correction to generate complete proofs (the prompts it employs are provided in full in Appendices~\ref{sec:appendix-prover-initial-prompt} and~\ref{sec:appendix-prover-subsequent-prompt}).

The agent receives the formal theorem statement along with appropriate \texttt{import} declarations and Mathlib~\citep{mathlib} dependencies. The model then generates a proof intended to close all goals. The system validates the proof through the Kimina server.

When verification fails, a corrector agent analyzes the error messages and generates a correction prompt that includes the original theorem statement, the failed proof attempt, and specific error messages from Lean. This self-correction loop continues until verification succeeds or the self-correction limits are reached (default: 2 attempts per pass, up to 32 passes).

For verified proofs, a parser agent requests the proof's AST from the Kimina server in order to, if required, use the AST in the reconstruction of an enclosing proof of which this proof may only be a part.

\subsection{Recursive Decomposition}

When direct proof generation fails, the system activates recursive decomposition, motivated by the POETRY algorithm~\citep{poetry} and the Hilbert~\citep{hilbert} system.

\textbf{Step 1: Theorem Retrieval.} The query generation agent employs a swappable LLM, which is by default Qwen 3 30GB~\citep{qwen2025}, to generate a number of natural language descriptions of theorems it thinks would be useful in the proof sketching task. (The prompts it employs are provided in full in Appendices~\ref{sec:appendix-search-query-initial} and \ref{sec:appendix-search-query-backtrack}). Using these natural language descriptions the theorem lookup agent then queries a vector database, we employ a LeanExplore variant~\citep{asher2025leanexploresearchenginelean}, to obtain a number of theorems that may be of use in the proof sketching task.

\textbf{Step 2: Proof Sketching.} The proof sketcher agent employs a swappable frontier LLM, which is by default GPT-5~\citep{openai2025gpt5}, along with the theorems retrieved in the previous step to generate a proof sketch that outlines the high-level structure using named hypotheses and placeholder proofs (the prompts it employs are provided in full in Appendices~\ref{sec:appendix-decomposer-initial-prompt},~\ref{sec:appendix-decomposer-subsequent-prompt}, and~\ref{sec:appendix-decomposer-backtrack-prompt}):

\begin{minted}{lean}
import Mathlib
import Aesop

set_option maxHeartbeats 0

open BigOperators Real Nat Topology Rat

variable (P Q R : Nat → Prop)

theorem complex_theorem (n : Nat) : P n := by
  have lemma1 : Q n := by sorry
  have lemma2 : Q n → P n := by sorry
  apply lemma2
  exact lemma1
\end{minted}

The sketch provides the overall proof strategy while deferring subgoal proofs to \texttt{sorry} tactics. The model is prompted to create sketches that are logically sound (in that the combination of assumed lemmas implies the theorem), self-contained (in that each lemma is independently provable), and appropriately granular (in that the lemmas are simpler than the original theorem but not trivial).

\textbf{Step 3: Sketch Validation.} The sketch checker submits the proof sketch to the Kimina server for validation. Lean verifies that the sketch is syntactically correct, that the types of all \texttt{have} statements are well-formed, and that the combination of assumed subgoals (with \texttt{sorry} proofs) logically implies the theorem statement.

When validation fails, a corrector agent analyzes the error messages and requests a sketch revision. Common errors include type mismatches in hypotheses and the \texttt{have} statements not entailing the main theorem. When repeated correction attempts fail (default: 6 attempts), a ``backtrack'' occurs and an ancestor is decomposed in a different manner, one that emphasizes alternative proof approaches.

\textbf{Step 4: AST-Based Extraction.} The parser agent requests the AST of the proof sketch from the Kimina server using the \texttt{/api/ast\_code} endpoint (the details of which appear in Appendix~\ref{sec:appendix-kimina}). The decomposition agent then traverses the AST in order to identify all unproven subgoals—that is, \texttt{have} statements followed by \texttt{by sorry}. For each subgoal, the system extracts the subgoal name, proof ``type'' (e.g. \texttt{sorry}), the complete theorem statement with appropriate context, and type information from the \texttt{sorries} metadata.

The AST traversal algorithm recursively walks the syntax tree, thereby identifying proof nodes and extracting subgoal information. A sketch of this algorithm takes the form:

\begin{minted}{python}
function ExtractSubgoals(ast):
    subgoals := []
    TraverseNode(ast, subgoals)
    return subgoals

function TraverseNode(node, subgoals):
    if node.type = "have" and node.proof.type = "sorry":
        subgoal := {name: node.name, type: node.type_expr}
        subgoals.append(subgoal)
    for each child in node.children:
        TraverseNode(child, subgoals)
\end{minted}

\textbf{Step 5: Recursive Proving.} Each extracted subgoal becomes a new formal theorem at depth $d+1$ in the proof tree. The system then recursively applies the complete proving pipeline (that is, direct proof generation with fallback to decomposition) to each subgoal. This creates a tree of proof obligations, wherein each node tracks its parent decomposition, sibling nodes represent independent subgoals, and depth limits serve to prevent infinite decomposition (default: 20 levels).

The recursive proving proceeds in breadth-first fashion: all subgoals at a given depth are processed before proceeding to deeper levels. When a subgoal exceeds the maximum depth limit (\texttt{max\_depth}, default: 20), the system attempts to backtrack to a backtrackable grandparent or higher ancestor in the proof tree. This ancestor is then re-decomposed with an alternative strategy, potentially yielding a shallower proof structure. Backtracking occurs only if the ancestor has not exhausted its self-correction attempts, thereby enabling the exploration of alternative decomposition strategies. If no backtrackable ancestor exists, the proof terminates with an appropriate error message. This mechanism serves to prevent premature termination when a shallower decomposition strategy might succeed, while still maintaining depth limits to prevent unbounded exploration.

\textbf{Step 6: Proof Reconstruction.} Once all subgoals verify successfully, the reconstruction algorithm (recursively) replaces each \texttt{sorry} placeholder in the parent sketch with the verified sub-proof body, producing a complete verified proof. A sketch of this algorithm takes the form:

\begin{minted}{python}
function ReconstructProof(node):
    if node is leaf:
        return node.formal_proof
    sketch := node.proof_sketch
    for each child in node.children:
        child_proof := ReconstructProof(child)
        proof_body := ExtractProofBody(child_proof)
        sketch := ReplaceSubgoal(sketch, child.name, proof_body)
    return sketch

function ExtractProofBody(proof):
    // Extract tactics after "by", excluding theorem declaration
    pattern := r"by\s+(.+)"
    return regex.match(pattern, proof).group(1)

function ReplaceSubgoal(sketch, name, proof_body):
    // Replace "have name : ... := by sorry" with complete proof
    pattern := f"have {name} : .* := by sorry"
    replacement := f"have {name} : ... := by {proof_body}"
    return regex.sub(pattern, replacement, sketch)
\end{minted}

The reconstruction maintains proper indentation and formatting.

\subsection{State Management}

The supervisor agent implements a state machine that determines the next action based upon the current proof state. The implementation employs LangGraph~\citep{langgraph}, which provides a framework for stateful, multi-agent applications with cyclic workflows.

The state manager tracks several categories of information: theorems awaiting formalization, validation, or semantic checking; theorems awaiting proof generation or verification; theorems requiring decomposition or sketch validation; the complete proof tree with parent-child relationships; verification status and error messages for all nodes; and the action history for debugging purposes. The state manager also implements depth-based backtracking: when children exceed the maximum depth, the system identifies backtrackable ancestors (at grandparent level or higher) and queues them for re-decomposition, rather than immediately terminating the proof attempt.

The supervisor queries the state manager in order to identify pending work, employing the following priority rules: formalize informal theorems (highest priority), validate formal theorems syntactically, check semantic equivalence, attempt direct proof generation, verify generated proofs, parse successful proofs into ASTs, query for useful theorems, decompose failed proofs, and recursively prove subgoals.

\section{Example}

To demonstrate the capabilities of the system, we consider the task of proving that the sum of two even numbers is even.

\textbf{Informal Input:} ``Prove that for any natural numbers $m$ and $n$, if $m$ is even and $n$ is even, then $m + n$ is even.''

\textbf{Formalization.} The Goedel-Formalizer-V2 model produces the following (examples include the system's standard Lean preamble):
\begin{minted}{lean}
theorem theorem_b2f45cfb951a : ∀ m n : ℕ, Even m → Even n → Even (m + n) := by
  sorry
\end{minted}

\textbf{Syntax Validation.} The Kimina server~\citep{kimina} confirms that the formalization is syntactically valid and type-correct.

\textbf{Semantic Checking.} qwen3:30b~\citep{qwen2025} verifies that the formalization preserves the meaning of the informal statement.

\textbf{Direct Proof Attempt.} Goedel-Prover-V2~\citep{goedel-prover-v2} using verifier-guided self-correction suggests the following proof:
\begin{minted}{lean}
import Mathlib
import Aesop

set_option maxHeartbeats 0

open BigOperators Real Nat Topology Rat

theorem theorem_b2f45cfb951a : ∀ m n : ℕ, Even m → Even n → Even (m + n) := by
  intro m n hm hn
  have h_main : Even (m + n) := by
    -- Extract the witnesses for m and n being even
    cases' hm with a ha
    cases' hn with b hb
    -- Rewrite m and n as 2a and 2b respectively
    rw
    -- Factor out the 2 from the sum 2a + 2b
    use a + b
    <;> ring
    <;> simp
    <;> ring
  -- The main result follows directly from the above steps
  exact h_main
\end{minted}

\textbf{Verification.} The Kimina server verifies that the proof succeeds.

For a more complex example that requires decomposition, we consider the task of proving the infinitude of primes. When direct proof generation fails after exhausting the configured self-correction loop (default: 2 attempts per pass, up to 32 passes), the system activates decomposition. The GPT-5 model generates the following:
\begin{minted}{lean}
import Mathlib
import Aesop

set_option maxHeartbeats 0

open BigOperators Real Nat Topology Rat

theorem infinitude_of_primes : ∀ n : Nat, ∃ p, p > n ∧ Prime p := by
  -- Step 1: Define product of all primes ≤ n
  have prod_primes_def :
      ∀ n, ∃ P, P = ∏ p in Finset.filter Prime (Finset.range (n + 1)), p := by
    sorry

  -- Step 2: For given n, take P = product of all primes ≤ n
  have choose_P :
      ∀ n, ∃ P, P = ∏ p in Finset.filter Prime (Finset.range (n + 1)), p := by
    sorry

  -- Step 3: Show that P + 1 has a prime divisor
  have prime_divisor_exists :
      ∀ n P, P = ∏ p in Finset.filter Prime (Finset.range (n + 1)), p →
        ∃ q, Prime q ∧ q ∣ (P + 1) := by
    sorry

  -- Step 4: Show that such a prime divisor q is greater than n
  have divisor_gt_n :
      ∀ n P q,
          P = ∏ p in Finset.filter Prime (Finset.range (n + 1)), p →
            Prime q → q ∣ (P + 1) → q > n := by
    sorry

  -- Step 5: Combine to conclude existence of a prime > n
  have conclusion : ∀ n, ∃ p, p > n ∧ Prime p := by
    sorry

  exact conclusion
\end{minted}

The AST parser identifies five subgoals: \texttt{prod\_primes\_def}, \texttt{choose\_P}, \texttt{prime\_divisor\_exists}, \texttt{divisor\_gt\_n}, and \texttt{conclusion}. Each subgoal is then independently proved (or, if necessary, further decomposed). Once all subgoals have been proven, the reconstruction algorithm substitutes their proofs into the sketch, thereby producing a complete verified proof.

\section{Implementation}

\subsection{Installation and Deployment}

The system is distributed through PyPI as \texttt{goedels-poetry}, thereby enabling straightforward installation with \texttt{pip install goedels-poetry}. The package installs the Python dependencies used by the framework (for example, LangGraph, LangChain, Kimina AST client, LeanExplore client, Rich, Typer) and a command-line interface accessible via \texttt{goedels\_poetry}. It does not install any model backends; callers supply their own OpenAI-compatible endpoints (e.g., OpenAI, Ollama, vLLM, or LM Studio) and run the required external services.

The system requires the Kimina Lean Server for verification. A packaged distribution, \texttt{kimina-ast-server} on PyPI, installs Lean 4, mathlib4, the AST exporter, and a FastAPI front end; running \texttt{kimina-ast-server setup} followed by \texttt{kimina-ast-server} starts the verifier (default: \texttt{http://localhost:8000}). The system also depends on the LeanExplore server (\texttt{lean-xplore} on PyPI) to provide semantic search over Mathlib declarations; starting \texttt{leanexplore http serve --backend local} (default: \texttt{http://localhost:8001/api/v1}) enables vector-database retrieval for the search and sketching agents. With both services running and an LLM backend configured, users may immediately begin proving theorems through either the CLI or the Python API.

For purposes of development and extension, the source code is available at \url{https://github.com/KellyJDavis/goedels-poetry} under the Apache 2.0 license. The modular architecture enables researchers to add custom agents, implement alternative decomposition strategies, or integrate additional LLM providers, all without modifying the core framework code.

\subsection{Architecture and Components}

The system is implemented in Python 3.10+ using LangGraph~\citep{langgraph} for multi-agent orchestration and LangChain~\citep{langchain} for LLM interaction. The codebase consists of approximately 5{,}000 lines organized into modular components: the Agents component (\texttt{goedels\_poetry/agents/}) implements 18 agent types spanning formalization, semantics checking, search-query generation, vector-database lookup, proving, verification, sketching, backtracking, and supervision; State Management (\texttt{goedels\_poetry/state.py}) provides tree-structured state representation and proof reconstruction logic (800 lines); AST Parsing (\texttt{goedels\_poetry/parsers/}) provides AST traversal and subgoal extraction utilities (now enriched with type binders recovered from Lean ``sorries''); Configuration (\texttt{goedels\_poetry/config/}) manages LLM selection and Kimina server connections. The decomposer uses an OpenAI backend (default: GPT-5) while the remaining agents default to a local Ollama endpoint; vLLM or LM Studio may be selected through configuration overrides.

The system supports flexible LLM configuration through a declarative INI file and environment variable overrides. Researchers may specify different models for each agent role, thereby enabling experimentation with various model combinations. 

Any OpenAI model compatible with the \texttt{langchain\_openai.ChatOpenAI} class is supported for theorem decomposition. Similarly, any model compatible with the \texttt{langchain\_openai.ChatOpenAI} class may be used for formalization, proving, semantic checking, and search-query generation. Vector retrieval is delegated to LeanExplore and is controlled via explicit configuration.

As an example, the default configuration (\texttt{goedels\_poetry/data/config.ini}) is:
\begin{minted}{ini}
[FORMALIZER_AGENT_LLM]
model = kdavis/goedel-formalizer-v2:32b
provider = ollama
url = http://localhost:11434/v1
api_key = ollama
max_tokens = 50000
num_ctx = 40960
max_retries = 10
max_remote_retries = 5

[PROVER_AGENT_LLM]
model = kdavis/Goedel-Prover-V2:32b
provider = ollama
url = http://localhost:11434/v1
api_key = ollama
max_tokens = 50000
num_ctx = 40960
max_self_correction_attempts = 2
max_depth = 20
max_pass = 32
max_remote_retries = 5

[SEMANTICS_AGENT_LLM]
model = qwen3:30b
provider = ollama
url = http://localhost:11434/v1
api_key = ollama
max_tokens = 50000
num_ctx = 262144
max_remote_retries = 5

[SEARCH_QUERY_AGENT_LLM]
model = qwen3:30b
provider = ollama
url = http://localhost:11434/v1
api_key = ollama
max_tokens = 50000
num_ctx = 262144
max_remote_retries = 5

[DECOMPOSER_AGENT_LLM]
model = gpt-5-2025-08-07
max_completion_tokens = 50000
max_remote_retries = 5
max_self_correction_attempts = 6

[KIMINA_LEAN_SERVER]
url = http://0.0.0.0:8000
max_retries = 5

[LEAN_EXPLORE_SERVER]
url = http://localhost:8001/api/v1
package_filters = Mathlib,Batteries,Std,Init,Lean
\end{minted}

This configuration specifies that formalization and proving employ locally hosted 32B models via Ollama, that semantic checking and search-query generation use qwen3:30b, that decomposition relies on an OpenAI GPT-5 endpoint, that verification is handled by the Kimina server, and that vector retrieval uses a local LeanExplore server filtered to Mathlib, Batteries, Std, Init, and Lean declarations. Providers may be switched to vLLM or LM Studio via environment variable overrides.

\subsection{Environment Variable Configuration}

All configuration parameters may be overridden through environment variables, thereby enabling deployment-specific customization without the need to modify configuration files or code. Environment variables employ the format \texttt{SECTION\_\_OPTION} (with double underscore separator, uppercase). For example:

\begin{minted}{bash}
# Override prover model
export PROVER_AGENT_LLM__MODEL="custom-model:latest"

# Use different Kimina server
export KIMINA_LEAN_SERVER__URL="http://production-server:8000"

# Adjust context window
export PROVER_AGENT_LLM__NUM_CTX="8192"
\end{minted}

This mechanism supports multiple deployment scenarios: development environments employing smaller models for faster iteration, continuous integration environments with dedicated verification servers, and production deployments with frontier models and distributed infrastructure. Environment variables take precedence over \texttt{config.ini} values, thereby enabling flexible configuration management without the need for reinstallation.

\subsection{Performance}

Without decomposition and using the default configuration, the system achieves a 90.4\% pass rate on miniF2F, as this is no more than the Goedel-Prover-V2 system using verifier-guided self-correction at pass@32. With RAG and recursive decomposition, this is significantly improved but has yet to be fully benchmarked.

The system's other performance characteristics are determined by several factors, with LLM inference forming the primary bottleneck. For informal theorems, formalization latency depends upon the model size and the complexity of the mathematical statement. Semantic checking with frontier models represents an additional overhead incurred per formalization attempt.

Direct proof generation with specialized models such as Goedel-Prover-V2 exhibits higher latency for complex theorems that require extensive context. Verification through the Kimina Lean Server is comparatively fast, as it involves deterministic type checking rather than generation. The system's retry mechanism serves to amplify latency when theorems fail direct proving, as multiple generation-verification cycles occur prior to decomposition.

Proof sketching with frontier models introduces additional latency that scales with both the theorem complexity and the number of decomposition strategies attempted. AST parsing latency depends primarily upon the complexity of import structures in the generated code, rather than upon the proof length.

The current implementation enforces consistent Lean preambles across all formal artifacts. Any incoming Lean theorem is first split into preamble and body components; the preamble is normalized, augmented with mandatory directives such as \texttt{set\_option maxHeartbeats 0}, and persisted with the proof state. Reconstruction routines subsequently recombine the stored preamble with regenerated proof bodies, ensuring that proof outputs carry the exact header required for downstream verification. The command-line interface mirrors this logic: it rejects formal inputs that omit a Lean header and emits diagnostic files explaining the requirement. This end-to-end normalization prevents subtle failures when default imports evolve and guarantees that proofs produced by the system are always accompanied by the proper Lean context.

Memory consumption is dominated by the proof tree structure, which retains all intermediate proofs, LLM conversation histories, and verification results. For deeply nested recursive decompositions, memory usage scales linearly with both tree depth and breadth. The configuration parameter \texttt{max\_depth} provides a safety limit intended to prevent unbounded memory growth.

The architecture naturally supports parallelization at the subgoal level, as sibling subgoals are independent and may be proved concurrently. The current implementation processes subgoals in parallel using LangGraph's \texttt{Send} API. This approach yields substantial speedups for theorems that decompose into many independent subgoals, with speedup limited primarily by the available computational resources and LLM API rate limits.

\subsection{Kimina Server Extensions}

As detailed in Appendix~\ref{sec:appendix-kimina}, we have extended the Kimina Lean Server with two AST endpoints: \texttt{POST /api/ast}, which exports ASTs for existing Lean modules (e.g., \texttt{Mathlib.Data.List.Basic}), and \texttt{POST /api/ast\_code}, which exports ASTs for dynamically provided Lean code snippets.

The \texttt{/api/ast\_code} endpoint proved essential for recursive decomposition. It accepts arbitrary Lean code, creates a temporary module, configures the correct import paths for Mathlib and the Lean standard library, invokes the AST exporter, and returns a structured AST representation. This enables real-time analysis of proof sketches generated by language models, all without the need for manual file management.

\section{Discussion}

\subsection{Comparison with POETRY}

While inspired by the work of POETRY~\citep{poetry}, our approach exhibits several notable differences. POETRY targets Isabelle~\citep{isabelle}, which employs a proof language and tactic system distinct from that of Lean 4. Lean's tactic mode encourages a more compositional style with explicit \texttt{have} statements, thereby simplifying subgoal extraction.

POETRY relies upon Isabelle's proof levels. In contrast, we employ explicit \texttt{have} statements with \texttt{sorry} placeholders, thereby making subgoals clearly identifiable within the AST. This explicitness serves to improve interpretability: proof sketches function as documentation of the proof strategy.

POETRY employs a single model for all tasks. We, in contrast, employ specialized models: Goedel-Prover-V2 for direct proving, and frontier LLMs for decomposition and semantic validation. This specialization serves to improve performance on each subtask. POETRY assumes that theorems are already formalized in Isabelle. We handle informal theorems through a dedicated formalization pipeline with semantic validation, thereby broadening the applicability of the approach.

\subsection{Concurrent Work: Hilbert}

During the development of our system, the Hilbert system~\citep{hilbert} independently arrived at similar conclusions regarding the utility of recursive decomposition for formal theorem proving. Both systems combine informal reasoning with formal verification, and both employ recursive strategies for complex theorems. Hilbert achieves impressive results: 99.2\% on miniF2F and 70.0\% on PutnamBench, thereby demonstrating the effectiveness of this approach.

Key architectural similarities include the recursive decomposition of complex problems, coordination between informal reasoning and formal verification, and the use of multi-component systems with specialized roles. In addition, like our work Hilbert employs a semantic theorem retriever for the purpose of finding theorems that assist in decomposition. Hilbert's implementation details and model configurations are not publicly available, whereas our system emphasizes open-source extensibility with straightforward model substitution. This convergence upon recursive decomposition and semantic theorem retrieval strategies from independent research efforts serves to validate this architectural direction for automated theorem proving.

\subsection{Limitations}

The system inherits the limitations of its constituent models. The default Goedel-Prover-V2 model excels on undergraduate-level mathematics but tends to struggle with specialized domains. Decomposition quality depends upon the default GPT-5 model's ability to identify productive proof strategies. For highly specialized domains (such as category theory or abstract algebra), the model may generate decompositions that prove unhelpful. That being said, all models may be swapped for more capable models. So these limitations may be exceeded.

The maximum recursion depth (default: 20) serves to prevent infinite decomposition. When subgoals exceed this depth, the system attempts to backtrack to earlier decomposition points and explore alternative strategies, thereby mitigating premature termination. However, theorems requiring very deep hierarchical reasoning may still fail if all backtrackable ancestors have exhausted their correction attempts or if no suitable alternative decomposition strategy exists. The system currently explores one decomposition path at a time.

Deep decomposition of complex theorems may require hundreds of API calls to frontier LLMs, thereby incurring significant cost. Techniques such as proof caching, early termination heuristics, and model distillation could serve to reduce these costs.

\subsection{Extensibility and Open Source Architecture}

The system is released as open source under the Apache 2.0 license, thereby enabling permissive use, modification, and extension. The codebase (\url{https://github.com/KellyJDavis/goedels-poetry}) has been designed for extensibility at multiple levels.

\textbf{Model Configuration.} The configuration system enables straightforward comparison of different LLMs for each agent role, all without the need for code modification. Researchers may evaluate whether larger models improve formalization accuracy, whether specialized models excel at specific proving tasks, or whether mixture-of-experts approaches outperform single-model strategies. The environment variable override mechanism supports A/B testing across different deployment configurations.

\textbf{Agent Extension.} The agent-based architecture supports experimentation with alternative reasoning strategies. New agents may be added through their implementation in LangChain and their registration with the system. This enables researchers to experiment with other strategies, combine multiple techniques, or use domain-specific proving tactics. The modular design ensures that new agents integrate seamlessly with the existing orchestration logic.

\textbf{Search Strategy Customization.} The tree structure enables alternative search strategies. The LangGraph framework can naturally support beam search (wherein multiple decomposition alternatives are maintained simultaneously), best-first search (wherein promising branches are prioritized based upon estimated difficulty), and adaptive backtracking (wherein unproductive branches are abandoned intelligently). Researchers may implement custom search heuristics without the need to modify too much of the core state management code.

\textbf{Training Data Generation.} Successful proof attempts can generate aligned (informal statement, formal theorem, verified proof) triples that prove suitable for fine-tuning. The hierarchical proofs provide additional training signal for the learning of decomposition strategies. The complete proof provenance enables synthetic data generation for the training of specialized formalization and proving models.

\textbf{Community Extensions.} The package distribution via PyPI, comprehensive documentation, and permissive license serve to facilitate community contributions. Researchers may publish custom agents, alternative decomposition strategies, or domain-specific configurations as separate packages that extend the core system. The modular architecture ensures that extensions remain compatible across version updates.

\section{Conclusion}

We have presented here a new approach to automated theorem proving, one that combines specialized language models with recursive decomposition. Through our extension of the Kimina Lean Server with AST parsing capabilities and our implementation of a multi-agent architecture that supports recursive proof strategies, we provide a practical system that serves to bridge informal mathematical reasoning with rigorous formal verification in Lean 4.

The key contributions of the system include a multi-stage formalization pipeline with semantic validation, AST-based subgoal extraction that enables systematic decomposition, a proof reconstruction algorithm that assembles verified sub-proofs into complete proofs, and a flexible configuration system that supports model substitution via INI files and environment variables.

The system is made available on PyPI as \texttt{goedels-poetry}, thereby enabling installation with \texttt{pip install goedels-poetry}. The open-source implementation (released under the Apache 2.0 license) at \url{https://github.com/KellyJDavis/goedels-poetry} facilitates experimentation with alternative models, decomposition strategies, and proof search algorithms. The modular architecture, built upon LangGraph and LangChain, serves to decouple orchestration logic from agent implementations, thereby enabling researchers to extend the system in diverse directions without the need to modify core functionality. Community contributions are encouraged through both the permissive licensing and the extensible architecture.

Possible future work includes the incorporation of alternative tree search techniques for exploring multiple decomposition paths, the integration of proof repair techniques~\citep{first2020baldur} for fixing near-correct proofs, the learning of decomposition strategies through reinforcement learning, and the extension of the system to collaborative theorem proving, wherein human experts guide decomposition strategies interactively.

\section*{Acknowledgments}

We gratefully acknowledge Project Numina for their development of the Kimina Lean Server, Justin Asher for his development of LeanExplore, the Lean community for their creation of the Lean 4 proof assistant and Mathlib library, and the developers of Goedel-Prover-V2, whose models power the core proving capabilities of this system. We also extend our thanks to the developers of LangGraph and LangChain for providing the infrastructure that enables this multi-agent architecture.

\bibliographystyle{plainnat}

\newpage
\appendix

\section{Kimina Lean Server AST Extensions}
\label{sec:appendix-kimina}

This appendix describes AST export functionality that was added to the Kimina Lean Server for the purpose of recursive decomposition.

\subsection{Motivation}

Recursive proof decomposition requires the programmatic extraction of unproven subgoals from proof sketches. While Lean's compiler identifies syntactic structure, the extraction of specific theorem statements with complete context and type information requires parsing of the abstract syntax tree. The Lean community's \texttt{ast-export} tool provides this capability, but the original Kimina server lacked an API endpoint through which to access it.

\subsection{AST Export Tool}

The \texttt{ast-export} tool (available at \url{https://github.com/KellyJDavis/ast_export}) exports Lean module definitions as structured JSON. For a given module, it produces a nested representation containing all theorem declarations (with their types and proof terms), hypothesis bindings within proofs (such as \texttt{have} statements), tactic invocations and their subgoals, and placeholders (\texttt{sorry}) that indicate unproven statements. The tool operates as a standalone Lake executable that is built against the same Lean version as the Kimina server.

\subsection{Implementation}

We added a router module \texttt{server/routers/ast.py} that implements two endpoints.

\subsubsection{Module AST Endpoint}

The \texttt{POST /api/ast} endpoint accepts a request that specifies modules, whether to export only the specified module (the \texttt{one} flag), and a timeout value. For each specified module, the endpoint invokes \texttt{lake exe ast-export} with the \texttt{--one} flag (which exports only the specified module, not its dependencies). The exporter writes JSON output to \texttt{.lake/build/lib/<module\_path>.out.json}. The endpoint then reads this file and returns structured results that include the module name, the AST, execution time, and diagnostics.

The endpoint employs a semaphore (\texttt{manager.ast\_semaphore}) to limit concurrent AST exports, thereby preventing resource exhaustion during batch processing.

\subsubsection{Code AST Endpoint}

The \texttt{POST /api/ast\_code} endpoint proved essential for the analysis of dynamically generated proof sketches. It accepts arbitrary Lean code, a module name, and a timeout value.

The endpoint creates a temporary directory with a \texttt{src/} subdirectory, and writes the provided code to \texttt{src/User/Code.lean}. It constructs \texttt{LEAN\_SRC\_PATH} to include the temporary \texttt{src/} directory, the mathlib4 project directory, and the \texttt{.lake/packages} directory. It constructs \texttt{LEAN\_PATH} to include compiled libraries (the \texttt{.olean} files), such that imported notations and macros are available during parsing. The endpoint then invokes \texttt{lake exe ast-export --one User.Code}, reads the generated \texttt{.lake/build/lib/User/}\allowbreak\texttt{Code.out.json}, returns the parsed AST, and performs cleanup.

The critical insight lay in the correct configuration of Lean source and library paths. Initially, the endpoint failed to resolve \texttt{import Mathlib} statements. The solution involved setting \texttt{LEAN\_SRC\_PATH} to include both the temporary source and the local mathlib4 checkout, setting \texttt{LEAN\_PATH} to include the compiled libraries, and running the exporter via \texttt{lake} within the AST exporter project directory. This configuration enables the AST exporter to process arbitrary Lean code with full Mathlib support.

\subsection{Error Handling}

Both endpoints implement robust error handling: asynchronous subprocess execution with timeout enforcement serves to prevent hanging on malformed input; Lean compilation errors are captured from stderr and returned in the error field; module names are validated against a regex in order to prevent directory traversal attacks; all AST export operations are logged with timing information; and responses include both execution time and AST size metrics.

\subsection{Integration}

The parser agents invoke the \texttt{/api/ast\_code} endpoint with proof sketches that have been generated by the decomposition agent. The returned AST is wrapped within an \texttt{AST} class (\texttt{goedels\_poetry/parsers/ast.py}) that provides methods for \texttt{get\_unproven\_subgoal\_names()} (which traverses the AST in order to identify all \texttt{sorry} placeholders and extract the names of associated hypotheses), \texttt{get\_named\_subgoal\_code(name)} (which retrieves complete Lean code for a specific subgoal, including context and type information from the \texttt{sorries} metadata), and \texttt{get\_ast()} (which returns the raw AST for custom analysis).

This abstraction serves to decouple decomposition logic from AST structure details, thereby enabling the maintenance of extraction logic independently of changes to the Lean AST format.

\subsection{Code Availability}

The AST endpoint implementation and AST exporter tool integration are available at 

\url{https://github.com/KellyJDavis/kimina-lean-server} 

\noindent in the \texttt{main} branch.

\section{Agent Prompts}
\label{sec:appendix-prompts}

This appendix provides the complete prompts used by each agent in the Gödel's Poetry system.

\subsection{Formalizer Agent Prompt}
\label{sec:appendix-formalizer-prompt}

The formalizer agent uses the following prompt to translate informal mathematical statements into Lean 4:

\begin{minted}{markdown}
Please autoformalize the following natural language problem statement in Lean 4. Use the following theorem name: {{ formal_statement_name }}

The natural language statement is:

{{ informal_statement }}

Think before you provide the lean statement.
\end{minted}

\subsection{Prover Agent Initial Prompt}
\label{sec:appendix-prover-initial-prompt}

The prover agent uses the following initial prompt for direct proof generation:

\begin{minted}{markdown}
Complete the following Lean 4 code:

```lean4
{{ formal_statement }}```

Before producing the Lean 4 code to formally prove the given theorem, provide a detailed proof plan outlining the main proof steps and strategies.
The plan should highlight key ideas, intermediate lemmas, and proof structures that will guide the construction of the final formal proof.

IMPORTANT: When you've completed your planning you should output your proof of the Lean 4 code within a ```lean4...``` code block.

IMPORTANT: Your proof should actually prove the theorem or lemma using Lean 4 code. It should not contain sorry, admit, or any other Lean 4 tactics that indicate an incomplete proof.

IMPORTANT: Your completion should contain ONLY the theorem/def and its proof. Do NOT include any import statements or preamble (like `import Mathlib`, `open`, `set_option`, `noncomputable section`, etc.) as these are already provided in the code above. Start directly with your theorem/definition/lemma declaration.
\end{minted}

\subsection{Prover Agent Subsequent Prompt}
\label{sec:appendix-prover-subsequent-prompt}

When proof attempts fail, the prover agent uses this correction prompt:

\begin{minted}{markdown}
The proof (Round {{ prev_round_num }}) is not correct. Following is the compilation error message, where we use <error></error> to signal the position of the error.

{{ error_message_for_prev_round }}

Before producing the Lean 4 code to formally prove the given theorem, provide a detailed analysis of the error message.

IMPORTANT: When you've completed your detailed analysis of the error message you should output your proof in Lean 4 code within a ```lean4...``` code block.

IMPORTANT: Your proof should actually prove the theorem or lemma using Lean 4 code. It should not contain sorry, admit, or any other Lean 4 tactics that indicate an incomplete proof.

IMPORTANT: Your corrected proof should contain ONLY the theorem/def and its proof. Do NOT include any import statements or preamble (like `import Mathlib`, `open`, `set_option`, `noncomputable section`, etc.). Start directly with your theorem/definition/lemma declaration.
\end{minted}

\subsection{Query Agent Initial Prompt}
\label{sec:appendix-search-query-initial}

The query agent initially uses the following prompt to generate natural language descriptions of potentially useful theorems:

\begin{minted}{markdown}
You are a search query generation assistant for a mathematical theorem proving system. Your task is to generate diverse search queries that will help find relevant theorems, lemmas, and tactics from a mathematical library (like Mathlib) that could be useful for proving the given formal theorem.
                                                                                                    
Given a Lean 4 formal theorem statement, generate 3-5 diverse search queries that cover different aspects:
- Key mathematical concepts and structures mentioned in the theorem                                 
- Related lemmas or theorems that might be useful                                                   
- Specific tactics or proof techniques that could apply                                             
- Important properties or relationships                                                             
                                                                                                    
Each query should be concise and focused on a specific aspect that could help in proving the theorem.
                                                                                                    
Here is the formal theorem statement:                                                               
                                                                                                    
```lean4                                                                                            
{{ formal_theorem }}                                                                                
```                                                                                                 
                                                                                                    
Generate your search queries below, each enclosed in <search> tags:                                 
                                                                                                    
<search>your first search query here</search>                                                       
<search>your second search query here</search>                                                      
<search>your third search query here</search>
\end{minted}

\subsection{Query Agent Backtrack Prompt}
\label{sec:appendix-search-query-backtrack}

The query agent when backtracking uses the following prompt to generate natural language descriptions of potentially useful theorems:

\begin{minted}{markdown}
You are a search query generation assistant for a mathematical theorem proving system. Your task is to generate diverse search queries that will help find relevant theorems, lemmas, and tactics from a mathematical library (like Mathlib) that could be useful for proving the given formal theorem.
                                                                                                    
**IMPORTANT**: A previous attempt to prove this theorem failed. The previous decomposition strategy was syntactically valid but the decomposed subgoals could not be proven after multiple attempts. This suggests the previous approach may not have been the right strategy.
                                                                                                    
Given the formal theorem statement and the conversation history about the previous failed attempt, generate 3-5 NEW and DIFFERENT search queries that:
- Explore alternative mathematical concepts or structures                                           
- Consider different proof techniques or approaches                                                 
- Focus on different aspects of the problem than the previous attempt                               
- May lead to a simpler or more direct proof path                                                   
                                                                                                    
Each query should be concise and focused on a specific aspect that could help in proving the theorem using a different strategy.
                                                                                                    
Here is the formal theorem statement:                                                               
                                                                                                    
```lean4                                                                                            
{{ formal_theorem }}                                                                                
```                                                                                                 
                                                                                                    
The conversation history above contains information about the previous failed attempt. Use that context to generate different search queries.
                                                                                                    
Generate your search queries below, each enclosed in <search> tags. Make sure they are different from what might have been used in the previous attempt:
                                                                                                    
<search>your first search query here</search>                                                       
<search>your second search query here</search>                                                      
<search>your third search query here</search>
\end{minted}

\subsection{Decomposer Agent Initial Prompt}
\label{sec:appendix-decomposer-initial-prompt}
The decomposer agent uses the following prompt for initial proof decomposition:

\begin{minted}{markdown}
You are a Lean 4 formal theorem prover assistant. Your task is to take a Lean theorem statement and:
                                                                                                    
1. Think and provide a natural-language proof sketch that explains the reasoning strategy.          
2. Decompose the proof into smaller formal Lean 4 have statements (subgoals) subordinate to the main Lean 4 theorem statement, i.e. contained within the by block of the main Lean 4 theorem.
3. When decomposing the proof, use existing proven theorems from the list below if warranted to help establish subgoals or simplify the decomposition.
4. Output Lean 4 code where each subgoal is expressed as a have statement ending with sorry, so that another prover can attempt to solve them recursively.
5. Ensure that the top-level Lean 4 theorem does not contain a sorry that is directly subordinate to it.
6. Ensure that the top-level Lean 4 theorem is entailed by the smaller formal Lean 4 have statements (subgoals) along, possibly, with other Lean 4 code within the by block of the main Lean 4 theorem.
7. Ensure that later subgoals may assume earlier ones as premises if helpful.                       
8. Do not attempt to fully solve the subgoals -- only set up the structured decomposition.
9. Do not introduce auxiliary lemmas or any other statements not subordinate to the main Lean 4 theorem.
10. In particular do not introduce an unexpected identifier or unexpected command such as "Complex" before the main Lean 4 theorem statement.
                                                                                                    
When decomposing the proof, you may find it helpful to use existing proven theorems from the list below. For example:
- You might use an existing theorem like `Nat.add_comm` to simplify expressions involving addition  
- You might apply a theorem about inequalities to establish relationships between terms             
- You might use a theorem about properties of functions to transform the goal into a more manageable form
- You might combine multiple existing theorems to build up to the desired conclusion                
                                                                                                    
Consider how these existing theorems might be incorporated into your proof sketch, either directly in the decomposition or as intermediate steps that help establish the subgoals.
                                                                                                    
{{ theorem_hints_section }}                                                                         
                                                                                                    
Example input theorem:                                                                              
                                                                                                    
```lean4                                                                                            
theorem induction_ineq_nsqlefactn (n : ℕ) (h : 4 ≤ n) : n ^ 2 ≤ n ! := by                          
  sorry```                                                                                          
                                                                                                    
Expected output format:                                                                             
                                                                                                    
* Natural language reasoning: high-level strategy (e.g., induction).                                
* Formal Lean code: theorem restated, followed by subgoal decomposition.                            
                                                                                                    
Example output (sketch):                                                                            
                                                                                                    
Natural language proof sketch:                                                                      
We prove by induction on n. Base case: verify for n = 4. Inductive step: assume k^2 ≤ k! for some k ≥ 4, then show (k+1)^2 ≤ (k+1)!.
                                                                                                    
Lean code with subgoals:                                                                            
```lean4                                                                                            
theorem induction_ineq_nsqlefactn (n : ℕ) (h : 4 ≤ n) : n ^ 2 ≤ n ! := by                          
  -- Base case                                                                                      
  have base_case : 4 ^ 2 ≤ 4 ! := by                                                                
    sorry                                                                                           
                                                                                                    
  -- Inductive step                                                                                 
  have inductive_step : ∀ k ≥ 4, k ^ 2 ≤ k ! → (k + 1) ^ 2 ≤ (k + 1) ! := by                        
    sorry                                                                                           
                                                                                                    
  -- Combine base case and inductive step                                                           
  have final_proof : ∀ n ≥ 4, n ^ 2 ≤ n ! := by                                                     
    sorry                                                                                           
```                                                                                                 
                                                                                                    
Here is the Lean 4 formal theorem to decompose:                                                     
                                                                                                    
```lean4                                                                                            
{{ formal_theorem }}```
\end{minted}

\subsection{Decomposer Agent Subsequent Prompt}
\label{sec:appendix-decomposer-subsequent-prompt}

When sketch validation fails, the decomposer agent uses this correction prompt:

\begin{minted}{markdown}
The proof sketch (Round {{ prev_round_num }}) is not correct. Following is the compilation error message, where we use <error></error> to signal the position of the error.

{{ error_message_for_prev_round }}

Before producing the Lean 4 code to sketch a proof to the given theorem, provide a detailed analysis of the error message.
\end{minted}

\subsection{Decomposer Agent Backtrack Prompt}
\label{sec:appendix-decomposer-backtrack-prompt}

When decomposition strategies fail repeatedly, the decomposer agent uses this backtrack prompt:

\begin{minted}{markdown}
The previous proof sketch (Round {{ prev_round_num }}) was syntactically valid, but the decomposed subgoals could not be proven after multiple attempts. This suggests that the decomposition strategy itself may not be the right approach for this theorem.
                                                                                                    
Please try a COMPLETELY DIFFERENT decomposition strategy. Consider:                                 
- A different proof technique (e.g., if the previous attempt used induction, try direct proof, contradiction, or case analysis)
- A different way to break down the problem into subgoals                                           
- Different intermediate lemmas that might be easier to prove                                       
- A simpler or more direct path to the conclusion                                                   
- Utilizing, if warranted, the possibly new theorems listed below in pursuit of your goal           
                                                                                                    
Before producing the new Lean 4 code to sketch a proof, provide:                                    
1. A brief analysis of why the previous decomposition might have been too difficult                 
2. A description of the new strategy you will try                                                   
3. An explanation of how this new approach differs from the previous one                            
                                                                                                    
Note that the theorems listed below may differ from those presented in previous attempts, as a new search strategy was used to find potentially useful theorems for this different decomposition approach. These theorems may or may not be new compared to previous attempts, but they represent theorems identified specifically for trying this alternative strategy. Consider how these theorems might be useful in your new decomposition approach.
                                                                                                    
{{ theorem_hints_section }}
\end{minted}

\subsection{Semantic Checker Agent Prompt}
\label{sec:appendix-semantic-prompt}

The semantic checker agent uses the following prompt to validate formalization accuracy:

\begin{minted}{markdown}
You will receive a math problem consisting of its natural language statement along with its formal statement in LEAN 4.

Please evaluate whether the formal LEAN statement appropriately translates the natural language statement based on the following criteria:

1. Key Elements: The problem's essential components are correctly represented in LEAN code.
2. Mathematical Accuracy: The translation preserves the accuracy of the mathematical content.
3. Structural Fidelity: The translation aligns closely with the original problem, maintaining its structure and purpose.
4. Comprehensiveness: All assumptions, conditions, and goals present in the natural language statement are included in the LEAN translation.

Your answer should be in the following format:

Thought: [Your Answer]

Judgement: [Your Answer, one of {Appropriate, Inappropriate}]

---

Following are the example problems label for the reasonability of their translation.

# Example 1:

## Original natural language statement of the problem:

For the graph of a certain quadratic $y = ax^2 + bx + c$, the vertex of the parabola is $(2,10)$, and one of the $x$-intercepts is $(1,0)$.  What is the $x$-coordinate of the other $x$-intercept?

## Translated formal statement:
```lean
theorem quadratic_other_intercept
    (f : $\mathbb{R}$ → $\mathbb{R}$)
    (a b c : $\mathbb{R}$)
    (h_quad : ∀ x, f x = a * x^2 + b * x + c)
    (h_vertex : f 2 = 10 ∧ ∀ x, f x ≤ f 2)  -- vertex at (2,10)
    (h_intercept1 : f 1 = 0) :               -- x-intercept at (1,0)
    f 3 = 0 := by                            -- other x-intercept at (3,0)
```

Thought: The Lean translation of the problem is appropriate because it accurately captures the intent and reasoning of the original problem. The problem's key elements, such as the vertex, axis of symmetry, and x-intercepts of the quadratic function, are correctly translated into Lean code. The logical flow of the proof mirrors the original reasoning, starting with the symmetry property of the quadratic function and using it to determine the second x-intercept. The mathematical accuracy is preserved, as the proof correctly applies the vertex property and symmetry to arrive at the solution. Furthermore, the translation aligns well with the original problem in natural language, maintaining fidelity to its structure and purpose. Overall, the translation is both faithful and complete, making it an appropriate representation of the original problem.

Judgement: Appropriate

# Example 2:

## Original natural language statement of the problem:

Draw a tangent line from the point $(4, 3)$ to the circle $(x-2)^2 + (y-1)^2 = 1$. What is the equation of the line that passes through the two tangent points?

## Translated formal statement:
```lean
theorem tangent_line_equation (x y : $\mathbb{R}$) :
  let P : $\mathbb{R}$ × $\mathbb{R}$ := (4, 3)  -- Point P
  let C : $\mathbb{R}$ × $\mathbb{R}$ := (2, 1)  -- Center of first circle
  let r : $\mathbb{R}$ := 1          -- Radius of first circle
  -- Points (x,y) satisfying both circle equations
  let on_first_circle := (x - 2)^2 + (y - 1)^2 = 1
  let on_second_circle := (x - 3)^2 + (y - 2)^2 = 2
  -- If point is on both circles
  on_first_circle ∧ on_second_circle →
  -- Then it lies on the line 2x + 2y - 7 = 0
  2*x + 2*y - 7 = 0 := by
```

Thought: The Lean translation of the problem is inappropriate because it fundamentally changes the intent of the original problem. The original problem asks to derive the equation of the tangent line passing through the intersection points of two circles, but the translation assumes the equation (2x + 2y - 7 = 0) is already given and instead asks to prove that the intersection points lie on this line. This shifts the problem from a construction task to a verification task, losing the original problem's focus on deriving the result through geometric and algebraic reasoning. Additionally, the translation omits the key reasoning step of subtracting the circle equations to derive the line equation, which is central to the original problem. As a result, the translation fails to accurately represent the problem's intent and educational value, making it an incomplete and inappropriate representation.

Judgement: Inappropriate


Example3:

## Original natural language statement of the problem:

If $a,b,c,d > 0$ and $abcd = 1$ , prove that

$ \frac{1}{a+b+c +1}+ \frac{1}{b+c+d+1}+\frac{1}{c+d+a+1}+\frac{1}{d+a+b+1} \le\frac{1}{a+3} +\frac{1}{b+3} + \frac{1}{c+3} + \frac{1}{d+3} $

## Translated formal statement:
```lean4
theorem lean_workbook_49553 (a b c d : $\mathbb{R}$) (habc : a * b * c * d = 1) : (1 / (a + b + c + 1) + 1 / (b + c + d + 1) + 1 / (c + d + a + 1) + 1 / (d + a + b + 1)) ≤ (1 / (a + 3) + 1 / (b + 3) + 1 / (c + 3) + 1 / (d + 3))  :=  by sorry
```

Thought: The Lean translation of the problem is inappropriate because the condition $a,b,c,d>0$ is ignored in the formal statement.

Judgement: Inappropriate



Example4:

## Original natural language statement of the problem:

If $a=b=c=2$ so $\sum_{cyc}\frac{(a-1)^2}{a^2+2}=\frac{1}{2}$ . We'll prove that $\frac{1}{2}$ is the answer.

## Translated formal statement:
```lean4
theorem lean_workbook_plus_1478 (a b c : $\mathbb{R}$) (ha : a = 2) (hb : b = 2) (hc : c = 2) : (a - 1) ^ 2 / (a ^ 2 + 2) + (b - 1) ^ 2 / (b ^ 2 + 2) + (c - 1) ^ 2 / (c ^ 2 + 2) = 1 / 2   :=  by sorry
```

Thought: The Lean translation of the problem is appropriate because it accurately captures the assumptions and the goal in the natural language statement.

Judgement: Appropriate

## Original natural language statement of the problem:

{{ informal_statement }}

## Translated formal statement:
```lean4
{{ formal_statement }}
```
\end{minted}

\end{document}